\documentclass[runningheads]{llncs}
\usepackage{graphicx}
\usepackage{comment}
\usepackage{adjustbox}
\usepackage{amsmath}
\usepackage{amssymb}
\usepackage{xcolor,colortbl}
\usepackage{float}

\begin{document}
\title{Gender Recognition in Informal and Formal Language Scenarios via Transfer Learning. \thanks{Supported by University of Antioquia.} }
\titlerunning{Gender Recognition in Text Data via Transfer Learning.}
%
\author{Daniel Escobar-Grisales\inst{1}\orcidID{0000-0002-3257-0134} \and
Juan Camilo Vásquez-Correa\inst{1,2,3}\orcidID{0000-0003-4946-9232} \and
Juan Rafael Orozco-Arroyave\inst{1,2}\orcidID{0000-0002-8507-0782}}
\authorrunning{D. Escobar-Grisales et al.}
%
\institute{GITA Lab. Faculty of Engineering, University of Antioquia UdeA, Medellín, Colombia. \and
Pattern Recognition Lab. Friedrich-Alexander-Universität Erlangen-Nürnberg, Germany \and Pratech Group, Medellín, Colombia.\\
\email{\{daniel.esobar,jcamilo.vasquez,rafael.orozco\}@udea.edu.co}}
\maketitle              
\begin{abstract}

The interest in demographic information retrieval based on text 
data has increased in the research community because applications 
have shown success in different sectors such as security, marketing,
heath-care, and others. 
Recognition and identification of demographic traits such as 
gender, age, location, or personality based on text data can 
help to improve different marketing strategies. 
For instance it makes it possible to segment and to personalize 
offers, thus products and services are exposed to the group of 
greatest interest.
This type of technology has been discussed widely in documents 
from social media. However, the methods have been poorly studied 
in data with a more formal structure, where there is no access 
to emoticons, mentions, and other linguistic phenomena that 
are only present in social media.
This paper proposes the use of recurrent and convolutional 
neural networks, and a transfer learning strategy for gender 
recognition in documents that are written in informal and 
formal languages. Models are tested in two different databases 
consisting of Tweets and call-center conversations. 
Accuracies of up to 75\% are achieved for both databases. 
The results also indicate that it is possible to transfer the 
knowledge from a system trained on a specific type of expressions or 
idioms such as those typically used in social media into a more 
formal type of text data, where the amount of data is more 
scarce and its structure is completely different.

\end{abstract}
\section{Introduction}

Demographic information retrieval consists in recognizing traits from 
a human being such as age, gender, personality, emotions, and others.
Typically the main aim is to create a user profile based on 
unstructured data. 
The retrieval of such information has different applications in 
forensics, security, sales, marketing, health-care, and many other sectors~\cite{hsieh2018author}. 
In e-commerce scenarios, this type of information provides 
advantages to companies in competitive environments because it 
allows to segment customers in order to offer personalized products and 
services which strengths their marketing 
strategies~\cite{dogan2019gender,hirt2019cognitive}. 
Although most of the demographic factors are explicitly collected 
through the registration process, this approach could be limited 
given that most of potential customers in online stores are 
anonymous. The automatic recognition of demographic variables 
such as gender can help to overcome these 
limitations~\cite{fernandez2018dimension}. 

Text data from customers can be obtained via transliterations of 
voice recordings, chats, surveys, and social media. 
These text resources can be processed to automatically recognize 
the gender of the users. 
Different studies have applied Natural Language Processing (NLP) 
techniques for gender recognition in text data, mainly from 
social media posts. In~\cite{markov2017language,martinc2017pan} the 
authors used Term Frequency-Inverse Document Frequency (TF-IDF) to extract features from tweets in the PAN17 corpus~\cite{rangel2017overview}, and reported accuracies for gender classification around 81\%. 
The authors in~\cite{basile2017n} used extracted features from 
TF-IDF as well as specific information only available in social 
media posts such as the frequency of female- and male-emojis. 
The authors reported an accuracy of 83.2\% in the 
PAN17~\cite{rangel2017overview} corpus for gender recognition.
Although the high accuracy reported in the study, the methodology 
would not be accurate to model text data written in more formal 
scenarios such as customer reviews, product surveys, opinion 
posts, and customer service chats, which have a different 
structure compared to the texts that can be found in social media data. 
In other study~\cite{li2017gender}, the authors proposed a system 
to classify the gender of the persons who wrote 100,000 posts from 
Weibo (Chinese social network similar to Tweeter). 
The system was based on a Word2Vec model, which achieved an 
accuracy of 62.9\%. The authors compared the performance of their 
model with human judgments, which accuracy was 60\%. 
This fact evidences that the problem of recognizing gender 
in written texts is very hard even for human readers. 
Wod2Vec models were also considered in~\cite{akhtyamova2017twitter} 
for gender recognition in the PAN17 corpus. 
The authors reported an accuracy of 69.5\% for the Tweets 
in Spanish.
There are some studies focused on gender classification using 
Deep Learning (DL) methods. However, when considering texts in 
Spanish, the number of studies is relatively 
 small~\cite{hsieh2018author,li2017gender}.
In~\cite{kodiyan2017author}, the authors proposed a methodology 
based on Bidirectional Gated Recurrent Units (GRUs) and an 
attention mechanism for gender classification in the PAN17 corpus. 
The authors worked with a Word2Vec model as input for their 
DL architecture and reported accuracies of up to 75.3\%.

According to the reviewed literature, gender classification based 
on text data has been mainly explored in social media scenarios,  
where the language is informal and the documents do not follow 
a formal structure~\cite{gonzalez2015analysis}. 
These types of documents use a number of language variants, 
styles, and other content like emojis that help to accurately 
recognize different demographic information.
There is a gap between models trained on formal and informal 
written language because a trained model with formal 
language data for a specific purpose will not achieve comparable 
results on an informal language scenario, or vice-versa~\cite{gu2020data}. 
Due to this reason, it is important to validate trained models for 
gender recognition in both types of languages: formal and informal. 
In addition, the recognition of demographic variables such as 
gender are under-explored in documents with a more formal 
structure.


This paper proposes a methodology based on Recurrent Neural Networks 
(RNNs) and Convolutional Neural Networks (CNNs) for gender 
recognition in informal and formal language scenarios. 
First, the models are trained and tested in the PAN17 corpus, 
which is a traditional dataset for gender classification in Tweets. 
The models originally trained using the PAN17 corpus are re-trained 
using a transfer learning strategy with data from call-center 
conversations, which are structured in a more formal language. 
Accuracies of up to 75\% are obtained, indicating that the 
proposed methodology is accurate for gender classification in 
documents written in formal and also in informal languages. 
Moreover, fine-tuned models using transfer learning show that 
despite the noise and lack of structure in documents written in 
informal language, they can be used to improve the accuracy of 
gender classification.

\section{Materials and Methods}
\subsection{Data}

\subsubsection{PAN17:}
We are particularly working with the Spanish data of the corpus, namely
PAN-CLEF 2017~\cite{rangel2017overview}. 
In this database, there are variants of Spanish from seven countries:
Argentina, Chile, Colombia, Mexico, Peru, Spain and Venezuela. 
The training set is composed by texts from 600 subjects from each country 
(300 female). Since each subject has 100 Tweets, there is a total of 
4200 subjects and 420000 Tweets in the dataset. 
The test set comprises data from 400 subjects from each country 
(200 female) for a total of 2800 subjects and 280000 Tweets.
For the sake of comparison with previous studies, we kept the 
original train and test sets. The training set was randomly divided 
into 80\% for training and 20\% to optimize the hyper-parameters of the models (development set). 
All data distribution was performed subject independent to avoid 
subject specific bias and to guarantee a better generalization 
capability of the models.

\subsubsection{Call Center Conversations:}

This corpus contains transliterations of conversations between customers 
and agents from a customer service center of a pension 
administration company in Colombia. 
Texts are manually generated by a group of linguistic experts based 
on the audio signals from the customers. Similarly, the label of the
gender is assigned based on the audio recordings processed by the 
linguists. 
Formal language is typically used by the customers when asking for a 
service, making a request, asking about certificates, and other
questions about the service provided by the company. 
This database comprises 220 transliterations of different 
customers (110 female). The average number of words for each 
conversation is 602, with a standard deviation of 554.

\subsection{Deep Learning Architectures for Gender Classification}

We consider two DL architectures in this case: an RNN with Bidirectional 
Long Short Term Memory (LSTM) cells, and a CNN with multiple temporal 
resolutions. These networks are trained with data from the 
PAN17 corpus. Then, a transfer learning strategy is applied 
to recognize gender from the call center conversations data. 

\subsubsection{Bidirectional Long Short Term Memory:} 

\begin{figure}[]
  \includegraphics[width=0.7\textwidth]{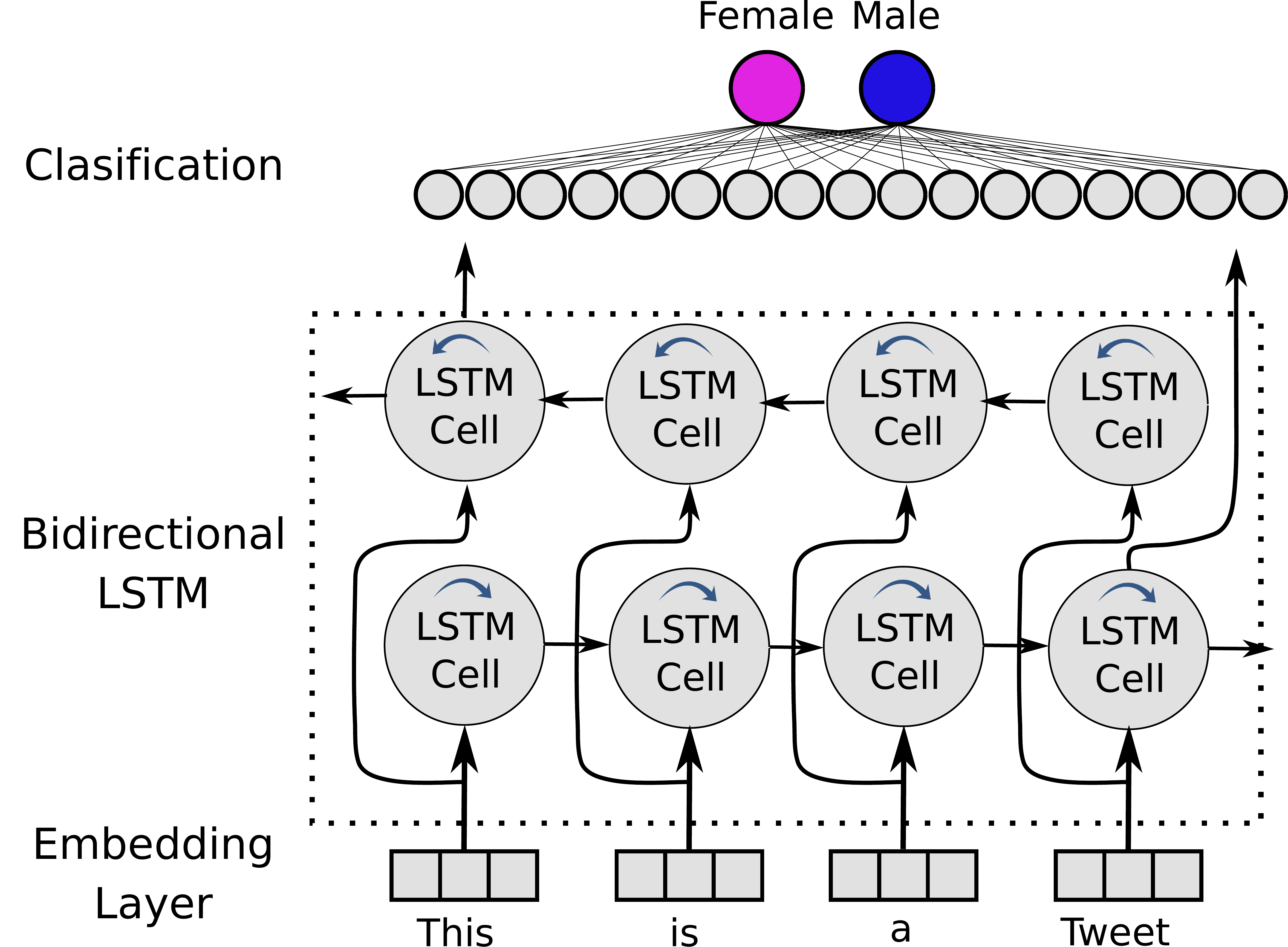}
\centering
\caption{Bi-LSTM architecture for gender classification in a Tweet.}
\label{fig:Bi-lstm}       
\end{figure}

The main idea of RNNs is to model a sequence of feature vectors 
based on the assumption that the output depends on the input 
features at the present time-step and on the output at the 
previous time-step. 
Conventional RNNs have a \emph{causal} structure, i.e., the output 
at the present time step only contains information from the past. 
However, many applications require information from the 
future~\cite{otter2020survey}. 
Bidirectional RNNs are created to address such a requirement by combing a 
layer that processes the input sequence forward through time with 
an additional layer that moves backwards the input sequence. 
Traditional RNNs also exhibit a vanishing gradient problem, which 
appears when modeling long temporal sequences. 
LSTM layers were proposed to solve this vanishing gradient problem 
by the inclusion of a \emph{long-term} memory to produce paths where 
the gradient can flow for long duration sequences such as sentences 
of a Tweet, or the ones that appear in a conversation with a
call-center agent~\cite{torfi2020natural}. 
We proposed the use of a Bidirectional LSTM (Bi-LSTM) network for our 
application. These architectures are widely used for different NLP 
tasks such as sentiment analysis in social media and product
reviews~\cite{arras2017explaining,minaee2019deep,trofimovich2016comparison}. 
A scheme of the considered architecture is shown in 
Fig.~\ref{fig:Bi-lstm}.

Words from the data are represented using a word-embedding layer.
The input to the Bi-LSTM layer consists of $k$ $d$-dimensional 
words-embedding vectors, where $k$ is the length of the sequence. 
The final decision about the gender of the subject is made at the 
output layer by using Softmax activation function.

\subsubsection{Convolutional Neural Network (CNN):}

\begin{figure}[]
  \includegraphics[width=0.7\textwidth]{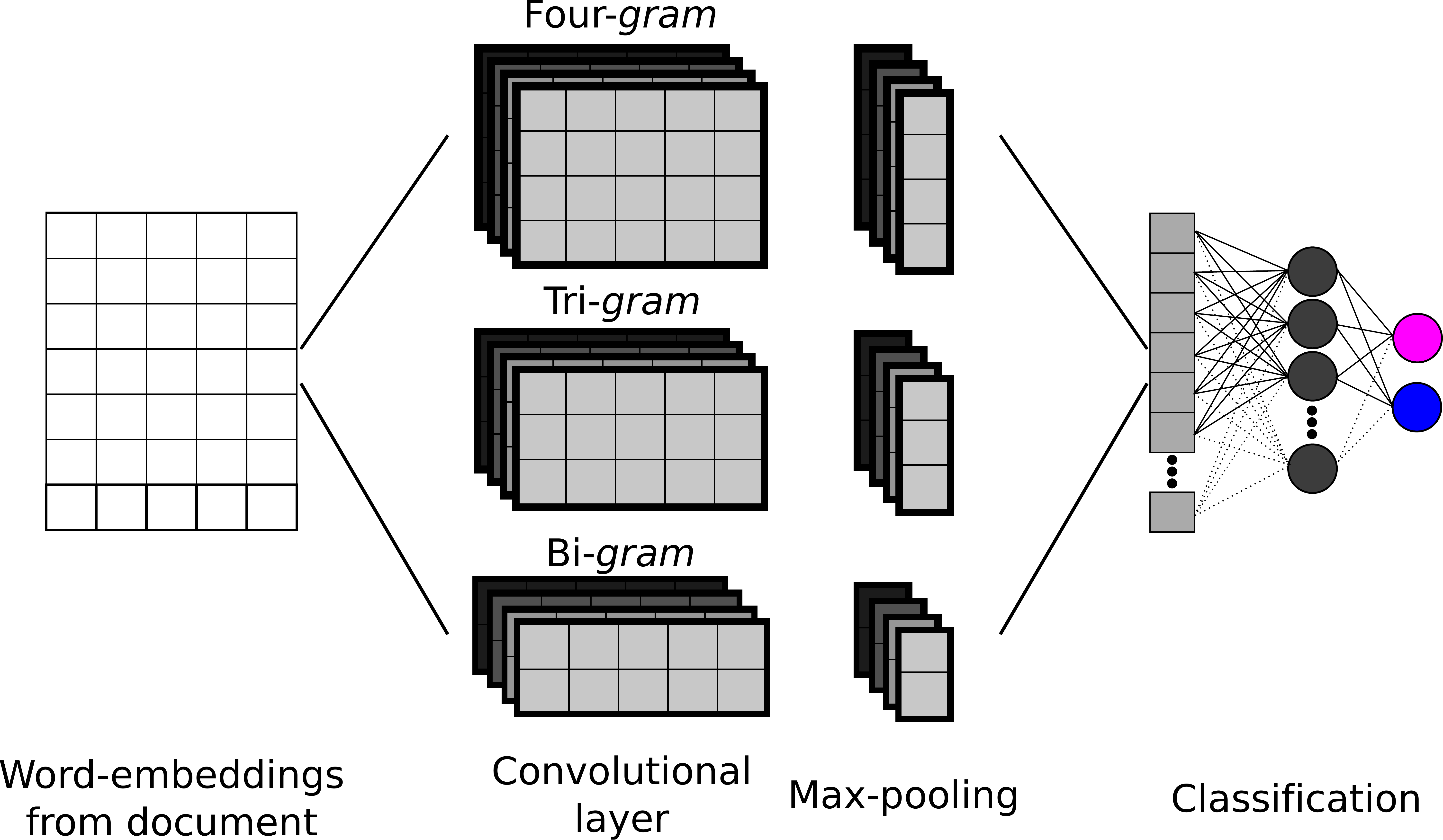}
\centering
\caption{CNN architecture for gender classification in a Tweet.}
\label{fig:CNN}       
\end{figure}

CNN-based architectures are designed to extract sentence representations 
by a composition of convolutional layers and a max-pooling operation over 
all resulting feature maps. We proposed the use of a parallel CNN 
architecture with different filter orders to exploit different temporal 
resolutions at the same time. Details of the architecture can be found in
Fig.~\ref{fig:CNN}. 
The output from the word-embedding layer is convolved with filters of 
different orders ($n$) and that correspond to different number of the $n$ 
in $n$-grams. The proposed CNN computes the convolution only in the temporal
dimension. After convolution, a max-pooling operation is applied. 
Finally, a fully connected layer is used for classification using a 
Softmax activation function.

\subsection{Training}

The networks considered in this work are implemented in Tensorflow 2.0, 
and are trained with a sparse categorical cross-entropy loss function 
using an Adam optimizer. 
An early stopping strategy is used to stop training when validation loss 
does not improve after 10 epochs. 
The embedding dimension $d$ is set to $100$. The vocabulary size for the 
tokenizer is set to $5000$ for the PAN17 corpus and $1500$ for the 
call-center conversations. The difference between the two vocabulary 
sizes is given because the number of unique words present in the training 
sets of each database. 
Hyper-parameters are optimized upon the validation accuracy and the 
simplest model.

\subsection{Transfer Learning}
We tested two approaches for the call-center conversations data: 
(1) training the network only using the data from the corresponding corpus, and 
(2) training the model via transfer learning by using a pre-trained model 
generated with the PAN17 corpus. 
For the transfer learning approach, the most accurate model for the 
PAN17 data is fine-tuned, but freezing the embedding layer in order 
to keep the tokenizer and a bigger vocabulary. Experiments without 
freezing the embedding layer were also performed but the results 
were not satisfactory.
The motivation for using transfer learning is to test whether the 
knowledge learned by a model trained with text data in informal 
language is useful to improve gender classification systems based on 
text with formal language, since it is generally common to collect  
large amounts of data with informal structure through social media, 
but it is difficult to collect written documents with a formal structure.

\section{Experiment and Results}

Two experiments are performed in this study. The first one consists in 
evaluating short sequences of texts, so the the architectures are trained 
and the gender of the subject is computed based on the average 
classification scores of all short texts from the same subject. 
Note that for PAN17 corpus, each tweet is a short text, while for 
call-center transliterations each conversation is divided into chunks 
with 60 words, similar to the proposed in~\cite{kodiyan2017author}. 
The second experiment consists in evaluating long assessment of texts. 
In this case, the complete text data from the subjects is entered
to the network at the same time. 
For the PAN17 corpus all Tweets are concatenated, and for the 
call-center conversations we consider the complete transliteration 
of each conversation. 
This strategy is only evaluated using the CNN-based approach because 
longer segments produced vanishing gradient problems in the 
Bi-LSTM network.

\subsubsection{Experiments with PAN17:} 

The results obtained for the PAN17 corpus (informal language) 
are shown in Table~\ref{tab:ResultsPAN17} for both approaches, 
short and long texts evaluation. Bi-LSTM and CNN networks are considered. 
The best results are obtained using the strategy with long texts
in the CNN. There is an improvement of up to 4\% in the accuracy
per subject with respect to the accuracy obtained with short texts. The
improvement in the F1-score is around 2\%.

\begin{table}[]
\centering
\caption{Results of the gender classification in the PAN-CLEF 2017 database.}
\label{tab:ResultsPAN17}  
\setlength{\tabcolsep}{3.5pt}
\begin{adjustbox}{max width=0.7\textwidth}
\begin{tabular}{lccc}
                   & \multicolumn{2}{c}{\textbf{Short texts}} & \textbf{Long texts} \\
                   & \textit{Bi-LSTM}    & \textit{CNN}    & \textit{CNN}                 \\ \hline
Accuracy per Tweet & 60.5                & 61.1            & -                   \\
Accuracy per subject         & 71.3                & 71.9            & \textbf{75.9}                \\
Precision          & 68.6                & 81.1            & \textbf{75.6}                \\
Recall             & 72.0                & 68.0            & \textbf{76.1}                \\
F1-Score           & 70.8                & 73.9            & \textbf{75.8}                \\ \hline
\end{tabular}
\end{adjustbox}
\end{table} 

\subsubsection{Experiments with Call-Center Conversations:} 

The results observed for the call center conversations 
(formal language) are shown in Table~\ref{tab:ResultsCallCenterTranscriptions}. 
The results include those obtained with and without applying transfer 
learning. 
The results also include the ones obtained using short and long texts. 
The results for this corpus are obtained following a 10-fold cross-validation 
strategy due to the small size of the corpus. 
The highest accuracy is obtained here also with the long texts, 
similar to the results obtained with the PAN17 corpus. 
In addition, note that the accuracy improves in up to 20\% when the 
transfer learning strategy is applied with respect to the accuracy 
obtained without using the pre-trained models. 
Note also that models using transfer learning show a smaller standard 
deviation which likely indicates that these methods are more stable.

\begin{table}[]
\centering
\caption{Results of the gender classification in the call-center conversations data. \textbf{TL}: transfer learning.}
\label{tab:ResultsCallCenterTranscriptions}  
\setlength{\tabcolsep}{3.5pt}
\begin{adjustbox}{max width=\textwidth}
\begin{tabular}{lcccc|cc}
                       & \multicolumn{4}{c}{\textbf{Short texts}}                                                                                                                                                                                                                                   & \multicolumn{2}{c}{\textbf{Long texts}}                                                                                       \\
                       & \textit{\begin{tabular}[c]{@{}c@{}}Bi-LSTM\\ Without TL\end{tabular}} & \textit{\begin{tabular}[c]{@{}c@{}}Bi-LSTM\\ With TL\end{tabular}} & \textit{\begin{tabular}[c]{@{}c@{}}CNN\\ Without TL\end{tabular}} & \textit{\begin{tabular}[c]{@{}c@{}}CNN\\ With TL\end{tabular}} & \textit{\begin{tabular}[c]{@{}c@{}}CNN\\ Without TL\end{tabular}} & \textit{\begin{tabular}[c]{@{}c@{}}CNN\\ With TL\end{tabular}} \\ \hline
Accuracy per text & 52.7 $\pm$ 6.43                                                           & 51.6 $\pm$ 5.07                                                        & 57.9 $\pm$ 9.20                                                       & 58.3 $\pm$ 6.48                                                    & -                                                                 & -                                                              \\
Accuracy per subject             & 54.2 $\pm$ 10.1                                                           & 56.4 $\pm$ 12.1                                                        & 65.9 $\pm$ 12.7                                                       & 62.9 $\pm$ 14.9                                                     & 55.9 $\pm$ 11.9                                                       & \textbf{75.0 $\pm$ 6.18}                                           \\
Precision              & 65.3 $\pm$ 29.2                                                           & 55.0 $\pm$ 13.8                                                        & 52.0 $\pm$ 22.2                                                       & 61.1 $\pm$ 17.9                                                    & 54.6 $\pm$ 25.4                                                       & \textbf{77.2 $\pm$ 8.12}                                           \\
Recall                 & 53.3 $\pm$ 22.7                                                           & 56.7 $\pm$ 13.2                                                        & 70.4 $\pm$ 17.2                                                       & 64.6 $\pm$ 15.2                                                    & 54.9 $\pm$ 34.8                                                       & \textbf{72.1 $\pm$ 10.4}                                           \\
F1-Score               & 55.2 $\pm$ 19.8                                                           & 55.2 $\pm$ 11.9                                                        & 57.8 $\pm$ 19.9                                                       & 61.1 $\pm$ 15.7                                                    & 48.5 $\pm$ 24.0                                                        & \textbf{73.8 $\pm$ 6.06}                                           \\ \hline
\end{tabular}
\end{adjustbox}
\end{table}

\section{Conclusions}

We proposed a methodology for automatic gender classification based on 
formal texts such as those available in social media posts, and based on
formal texts collected in call center conversations. 
Different deep learning models are evaluated including one on 
Bi-LSTMs, another one based on CNNs and a transfer learning approach, which
is pre-trained with data collected from social networks.
The transfer learning method is fine-tuned to improve the accuracy of the 
model designed for text classification in formal languages. 
The results indicate that it is possible to classify the gender of a 
person based on his/her written texts with accuracies of about 75\% 
in informal and formal language scenarios. 
The use of a transfer learning strategy improved the accuracy in 
scenarios where it is more difficult to collect data like in call-center
conversations, indicating that this strategy is suitable for companies or
sectors where it is not possible to create large datasets from scratch. 
The models using transfer learning are also more stable and 
generalize better than others where the neural networks are trained from 
scratch. This is very positive since it is possible to benefit from 
the large amounts of text data that are available in other domains like 
the social networks.
The proposed approaches can be extended to other applications related to 
demographic information retrieval such as age recognition, geographic 
location, personality of the subjects, and others, which would allow 
the building of more complete and specific author/customer profiles. 

\section*{Acknowledgments}
This work was funded by the company Pratech Group S.A.S and the University of Antioquia, grant \# PI2019-24110. We would like to thank the Natural Language Engineering Laboratory of the Universidad Politécnica de Valencia for providing access to one of the the databases used in this work.


\bibliographystyle{splncs04}
\bibliography{bibliography}
\end{document}